\documentclass[journal]{IEEEtran}
\usepackage{cite}

\ifCLASSINFOpdf
  \usepackage[pdftex]{graphicx}
\else
\fi
\usepackage{amsmath}
\usepackage{array}

\ifCLASSOPTIONcompsoc
  \usepackage[caption=false,font=normalsize,labelfont=sf,textfont=sf]{subfig}
\else
  \usepackage[caption=false,font=footnotesize]{subfig}
\fi
\usepackage{url}

\usepackage{bbm}

\begin{document}
%
\title{Vehicular Multi-Object Tracking with Persistent Detector Failures}
%
%
%

\author{Michael~Motro and Joydeep Ghosh
\thanks{Authors are with the Department
of Electrical and Computer Engineering, University of Texas, Austin,
TX, 78701 USA.}
\thanks{This work was supported by Qualcomm Research.}
}

%
%

\markboth{submit to IEEE Transactions IV, September 2019}
{Shell \MakeLowercase{\textit{et al.}}: Bare Demo of IEEEtran.cls for IEEE Journals}
%



\maketitle

\begin{abstract}
Autonomous vehicles often perceive the environment by feeding sensor data to a learned detector algorithm, then feeding detections to a multi-object tracker that models object motions over time. Probabilistic models of multi-object trackers typically assume that errors in the detector algorithm occur randomly over time. We instead assume that undetected objects and false detections will persist in certain conditions, and modify the tracking framework to account for them. The modifications are tested on a vehicle tracking dataset using a state-of-the-art lidar-based detector, a novel lightweight detector, and a fusion of camera and lidar detectors. For each detector, the persistence modifications notably improve performance and enable the model to outperform baseline trackers.
\end{abstract}

\begin{IEEEkeywords}
multi-object tracking (MOT), object detection, intelligent vehicles
\end{IEEEkeywords}

%
\IEEEpeerreviewmaketitle

\section{Introduction}
%
%
%
%
\IEEEPARstart{A}{mbitious} applications of intelligent vehicles have demanded new paradigms of perception. For instance, self-driving vehicles in urban settings may need to detect several types of stationary and moving objects within tens of meters in all directions, and understand these objects' motion. Camera and laser-based object detection is frequently performed by learned algorithms that directly transform raw data into object estimates. Current sensors and detectors are not capable of perfectly detecting all objects in all scenarios. They can err in three ways: by inaccurately reporting the position and characteristics of an object, by not detecting a present object, and by reporting a detection that does not correspond to a true object. We refer to the latter two as false negative and false positive detector failures. The quality of detection is proportional to the visibility of the object, the resolution of the sensor, and the computational resources available.

Multi-object tracking (MOT) uses detections at each timestep to maintain an estimate of all relevant objects and their past motion. Single-object trackers are often built around a probabilistic model that includes known characteristics of object motion and sensor behavior. The same approach can be applied to MOT by propagating a joint distribution across the states of all objects. The probabilistic approach to MOT also specifies a probability that each tracked object is actually present, which helps when tracking in the presence of detector failures.

However, several aspects of intelligent vehicle perception challenge the probabilistic implementation of MOT. Inter-object dependencies, such as the constraint that objects cannot overlap in space, complicate the multi-object distribution. Detection quality may depend on the environment, for instance if obstacles block line-of-sight. The position of the sensor itself may be uncertain if the robot is moving in an unknown environment, or if its sensors are moveable or damaged. Finally, tracking is typically formulated with the assumption that errors in object detection and localization are distributed randomly and independently over time. When detection is being performed by complex but powerful approaches such as machine learning algorithms, errors are more likely to be consistent functions of the sensor and environment. 

The inaccuracies of certain kinds of detector may be frequent or persistent enough to limit the benefits of tracking. The opposite can also hold: for highly accurate detectors on limited tasks, simple tracking techniques have sometimes been shown to outperform more complex trackers built around weaker detectors \cite{intro_lazytracking, fakes_simpletracking2, simpletracking1}. Thus a common approach is to put the majority of research effort and implementation cost into high-quality detectors, despite the cost in sensor and computational resources.


Some practical properties of vehicular object detectors have been considered in multi-object tracking. For instance, line-of-sight occlusion \cite{me_mwo} and localization \cite{localizationandtracking} can be accounted for with straightforward modifications to tracking algorithms. We address the temporal persistence of detector failures. We provide simple examples to suggest that failure to address this detector characteristic will cause trackers to underperform. We also derive altered forms of the standard probabilistic multi-object tracker for persistent false negative detections and false positive detections, by augmenting the state of tracked objects. The resulting tracker is applied to vehicle tracking on the Kitti dataset. The method is not detector-specific, and is tested with three different detectors: a state-of-the-art deep network for high-resolution lidar, a novel lidar-based detector that is more interpretable and doesn't require a graphics processing unit, and a fusion of a monocular detector with sparse lidar data.

\section{Vehicular Object Detection} \label{section_fakesobjdet}
Three types of sensors are widely used in autonomous vehicle perception. Radar is often processed in two stages. The low-level stage performs frequency-based processing of received waves and returns points in space expected to be filled by an object. The high-level stage clusters points into distinct objects and false positives \cite{mwo_variationalradar}. Machine learning for radar detection is less common in the literature, possibly because less unprocessed radar data is publicly available \cite{data_recentsurvey}. Lidar returns were traditionally handled as a set of horizontal line segments or curves, or as an unordered set of points in 3D space termed a point cloud. Recent research on lidar perception focuses instead on learned algorithms. Camera information is difficult to apply to most perception tasks with simple handmade rules, and learned algorithms have been utilized for decades. We focus on learned algorithms for the task of object detection.

An object detection algorithm takes the input from a sensor or sensors of a multi-object environment and outputs a set of objects. Depending on the environment and sensor, this task may be highly separable - that is, the sensory input that provides information about one object is largely irrelevant to the other objects. Additionally, highly similar objects are often unlikely to exist - for instance, solid 3D objects cannot exist in overlapping areas. These characteristics motivate the classification-suppression approach to object detection. A classification model is trained to provide a score indicating the likelihood that a specific region in the environment, termed an \textit{anchor}, contains an object. During the classification step, this model is applied to many overlapping anchors. Then, in the non-maximum suppression step, only the highest-scoring of overlapping anchors is maintained as a detected object.


False negative errors in object detection occur when the sensor does not provide information about an object, or when the detection algorithm fails to use sensor information to correctly classify an object. False negatives cannot be `fixed' by a tracker, but they can be characterized so that tracked objects are handled correctly while undetected. Object detection algorithm failures are difficult to characterize, especially for complex black box algorithms. However, it is reasonable to assume that inaccuracies in the detector are consistent for highly similar inputs, for instance if the same object is viewed twice in identical surroundings. False positive errors are detections that do not correspond to an actual object. For vehicular applications, it is reasonable to assume that most false positives are stationary objects - or more directly, that most large moving objects should be tracked and detected. False positive detections therefore occur when the stationary features of a location are viewed from a certain perspective.

\subsubsection{Image Object Detection}
Pedestrian detection in images is one of the oldest applications of learned object detectors, with research still ongoing \cite{data_mot}. Until 2015, the most popular approaches were variants of the Viola Jones detector \cite{fakes_violajones}. This method convolutionally generates features across the image, then trains a boosted classifier. Boosted classifiers are a set of simple models, each of which returns a classification score. A few models can usually accurately classify most inputs, a property termed attentional cascade in computer vision \cite{fakes_violajones}. Hence the model combines high precision power for difficult inputs with high speed for the average input, making it particularly suited to object detection. However, the model was not generally found powerful enough to classify on raw pixel data, so engineered multipixel features were generated first. 

Deep convolutional neural networks (CNNs) have become the most popular method for image-based object detection and image or lidar-based traffic scene perception. They still follow the classification-suppression approach but have several clear advantages. Their deep structure allows for direct learning of low-level features, which have been shown to outperform similar handmade features \cite{detect_pointpillars}. Additionally, neural networks are amenable to being trained and utilized with GPUs. The disadvantages of deep networks are a purported brittleness to structure and hyperparameter choice, and slow inference due to model complexity. Most camera-based detectors aim to locate objects' positions within the image, which is termed 2D detection. Camera-based 3D object detectors \cite{monogrnet, traffic_monopmbm} additionally estimate an object's orientation and distance from the camera. Current state-of-the-art camera-based 3D vehicle detectors can reliably estimate distance within several meters.


\subsubsection{Lidar Object Detection}
While the term lidar applies to any laser-based ranging sensor, we focus on the rotating multi-laser lidars that are commonly used for wide-view or 360-view perception of traffic environments. These sensors return a set of 3D points that follow sparse vertical lines. In early research on autonomous driving, object detection from laser scanners was performed by clustering points into distinct objects and separately determining the identity of each object based on its points. Classifiers based on engineered features \cite{fakes_shallowlidarclassify, fakes_shallowlidarclass2} or even handmade models \cite{fakes_kampker, fakes_shallowlidarrules2} can identify vehicles or pedestrians in the immediate vicinity. 
However, as more ambitious driving capabilities were explored, the goal of perception shifted to include a variety of stationary and moving objects within a broader radius of the car. The first public benchmark of 3D object detection was established on the Kitti dataset \cite{data_kitti} in 2017. To date, all of the methods on the public leaderboard are deep CNN object detectors save for \cite{fakes_pydriver}.
The majority of high-performing methods are lidar-based or lidar-and-image-based CNNs utilizing GPUs. Lidar-based object detection seems to require specialized network implementations to operate more than 10 times a second on a GPU - for instance by sparsifying the convolutional operations \cite{detect_pointpillars, fakes_second}.
\section{Tracking}
This section presents the standard probabilistic formulation of multi-object tracking with a single object detector. We use random set notation in order to write a complete mathematical statement for the standard tracking model, which will then be modified. We do not go through the full derivation of a specific tracker, in particular avoiding the data association step as its implementation is largely unaffected by the discussed subject.

\subsection{Single-object Tracking}
Each tracked object is parameterized with a random state $x$. This includes observable features such as the object's location, shape, and class, as well as latent features such as the object's motion. The object is estimated by propagating a distribution $P(x)$. A motion model $P(x^+|x)$ is assumed, which describes the transformation of the object state after a certain period of time. The prediction step of a single-object tracker uses the prior estimate and the motion model to calculate $P(x^+)$. Each detection has a fixed set of values $z$. In the case that a single object is detected, the detection distribution $P(z|x)$ is assumed. Given an object distribution and a detection, as well the assumption that the detection corresponds to this object, the object's distribution can be updated with a simple application of Bayes rule $P(x|z) = \frac{P(x)P(z|x)}{\int_x P(x)P(z|x)}$. Common probability distributions assumed in tracking models are Gaussians, particle mixtures, and mixtures of Gaussians.

\subsection{Multi-object Tracking}
\label{subsec_backmot}
Probabilistic multi-object tracking estimates probability distributions over sets of estimated object states $X=\{x_1, \cdots, x_{|X|}\}$. Random sets differ from vector-valued random variables in that they have no intrinsic ordering and their cardinality $|X|$ is also random. A distribution on a set can be written in terms of a cardinality distribution and a set of joint distributions (one for each cardinality).
\begin{equation}
P(X) = P(|X|=K) \sum_{k\leftrightarrow i}{P\left( x_{k:k\rightarrow 1},x_{k:k\rightarrow 2}, \ldots ,x_{k:k\rightarrow K} \right) }
\end{equation}
Where all possible one-to-one mappings of indices $k \in \{ 1,\ldots ,K\}, i \in \{ 1,\ldots ,K\}$ are summed. For multi-object tracking, we consider a distribution of the form
\begin{align} \label{eq_multiplehypothesisdistribution}
P(X) = e^{-\kappa_U} \sum_{k\leftrightarrow i} R(k\leftrightarrow i) \prod_{k\rightarrow i}p_i(x_k) \prod_{k\rightarrow \emptyset} \kappa P_U (x_k)
\end{align}
This distribution maintains some number of independent, single-object component distributions indexed by $i$. In this case, the mapping from realized objects $k$ to distribution components $i$ is not one-to-one, as some components are not matched to true objects and vice versa. The null index $\emptyset$ is used for indices with no match. Additionally, a state distribution $P_U (x)$ and rate parameter $\kappa_U$ is maintained to specify regions in which an unknown number of objects may exist. In practice, these are objects that have recently entered the sensor view and have not yet been detected, or objects with too uncertain a position to be tracked accurately. The term $R(k\leftrightarrow i)$ specifies the joint probability of the components existing. A broad variety of trackers can be specified by an appropriate choice of $R(k\leftrightarrow i)$. For instance, the Poisson Multi-Bernoulli Mixture (PMBM) can be specified as
\begin{equation}
    R(k\leftrightarrow i) = \sum_H w_H \prod_{i:k\rightarrow i} r_{H,i} \prod_{i:\emptyset\rightarrow i} (1-r_{H,i})
\end{equation}
And this tracker is easily convertible other trackers such as the standard multi-hypothesis tracker, the GLMB, the track-oriented Multi-Bernoulli filter, and others \cite{tracking_pmbm}.

\subsubsection{Standard Multi-object Motion Model} \label{subsec_smotm}
A multi-object motion model specifies the transformation of the set of objects $X$ into $X^+$ after a period of time. In addition to a motion model for individual objects $P(x^+|x)$, the entry and exit of objects from the relevant area must be modeled. This is typically achieved by specifying a probability of object survival $r_S(x)$, and a distribution and rate for new objects $P_{\text{new}}(x), \kappa_{\text{new}}$. The model is directly written as
\begin{align}
\label{eq_smotm}
    P(X^+|X) = e^{-\kappa_{\text{new}}} \sum_{k\leftrightarrow i} \prod_{k\rightarrow i} r_S(x_i)P(&x_k^+|x_i)  \prod_{\emptyset\rightarrow i} \left(1-r_S(x_i)\right) \nonumber \\ &\prod_{k\rightarrow \emptyset} \kappa_{\text{new}} P_{\text{new}} (x_k^+)
\end{align}
and the predicted estimates are
\begin{align}
&P_i(x^+) = \frac{\int_x P_i(x) r_S(x) P(x^+|x)}{r_{S,i}} \label{eq_sopredict} \\
&\kappa^+_U = \kappa_U + \kappa_{\text{new}} \\
&P_U(x^+) = \frac{\kappa_U}{\kappa^+_U}\int_x P_U(x)r_S(x)P(x^+|x) + \frac{\kappa_{\text{new}}}{\kappa^+_U} P_{\text{new}}(x^+) \\
&R^+(k\leftrightarrow i) = R(k\leftrightarrow i) \prod_{k\rightarrow i} r_{S,i} \prod_{\emptyset\rightarrow i} (1-r_{S,i}) \\
&r_{S,i} = \int_x r_S(x) P_i(x) \nonumber
\end{align}

\subsubsection{Standard Multi-object Measurement Model} \label{subsec_smm}
A multi-object measurement model specifies the probability distribution of a set of detections $Z = \left\{ z_1, \cdots, z_{|Z|}\right\}$ given a current set of objects $X$. The standard model assumes that each object generates at most one measurement. Additionally, the probability that an object generates a measurement is independent of other objects, and the values of a generated measurement are independent of objects that did not generate it. Finally, some measurements may be erroneous and not correspond to any object. These are termed false positives, and are assumed to be generated by a Poisson process. $r_D$ refers to the probability that an object generates some detection, while $P(z|x)$ is the distribution over measurement $z$ given that it was generated by object $x$. $P_F$ and $\kappa_F$ specify the distribution over the set of false measurements.
\begin{align} \label{eq_smmm}
P(Z|X) = e^{-\kappa_F} \sum_{k\leftrightarrow j} \prod_{k\rightarrow j}r_D(x_k) P(z_j|&x_k) \prod_{k\rightarrow \emptyset}{(1-r_D(x_k))} \nonumber \\
&\prod_{\emptyset\rightarrow j} \kappa_F P_F(z_j)
\end{align}
Alternative measurement models consider detectors such that each object has many measurements \cite{mwo_multimeasurement} or handle special cases of interdependence between objects and measurements. The multi-object tracking update step calculates the posterior probability $P(X|Z)$. The expressions regarding each object can be separated into likelihood constants and posterior distributions.
\begin{align}
&P_{i,j}(z_j) = \int_{x} P_i(x)r_D(x)P(z_j|x) \label{eq_matchlik}\\
&P_{i,\emptyset} = \int_{x} P_i(x) (1-r_D(x)) \label{eq_objmisslik} \\
&P_{\emptyset,j}(z_j) = \kappa_F P_F(z_j) + \kappa_U \int_{x} P_U(x)r_D(x)P(z_j|x) \label{eq_msmtmisslik} \\
&P_{\emptyset,\emptyset} = \kappa_U \int_{x} P_U(x)(1-r_D(x)) \label{eq_sproutlik} \\
&P_{i,j}(x|z_j) = \frac{P_i(x)r_D(x)P(z_j|x)}{P_{i,j}(z_j)} \label{eq_matchpost} \\
&P_{i,\emptyset}(x) = \frac{P_i(x) (1-r_D(x))}{P_{i,\emptyset}} \label{eq_objmisspost}\\
&P_{\emptyset,j}(x|z_j) = \frac{P_U(x)r_D(x)P(z_j|x)}{\int_{x} P_U(x)r_D(x)P(z_j|x)} \label{eq_msmtmisspost}\\
&P_{\emptyset,\emptyset}(x) = \frac{P_U(x)(1-r_D(x))}{\int_{x} P_U(x)(1-r_D(x))} \label{eq_sproutpost}
\end{align}
We collectively refer to these as the multi-object tracking update expressions. Note that they are similar to the measurement likelihood and object update expressions for single-object tracking. The full posterior distribution can then be written up to some normalizing constant.
\begin{align} \label{eq_motposterior}
P(X|Z) = \sum_{k\leftrightarrow i,j} &R_+(k\leftrightarrow i,j) \prod_{k\rightarrow i,j} P_{i,j}(x_k|z_j) \prod_{k\rightarrow i,\emptyset} P_{i,\emptyset}(x_k) \nonumber \\
&\prod_{k\rightarrow \emptyset,j} P_{\emptyset,j}(x_k|z_j) \prod_{k\rightarrow \emptyset,\emptyset} P_{\emptyset,\emptyset} P_{\emptyset,\emptyset}(x_k) \\
R_+(k\leftrightarrow i,j) \propto R&(k\leftrightarrow i) A(k\leftrightarrow i,j) \prod_{k\rightarrow i,j} P_{i,j}(z_j) \nonumber \\
&\prod_{k\rightarrow i,\emptyset} P_{i,\emptyset} \prod_{k\rightarrow \emptyset,j} P_{\emptyset,j}(z_j) \prod_{\emptyset\rightarrow j} \kappa_F P_F(z_j)
\end{align}
This form is equivalent to the initial distribution in \eqref{eq_multiplehypothesisdistribution}, as is necessary for sequential updates in a tracker. The posterior distribution has one potential object for every matching of prior object $i$ and measurement $j$, as well as potential objects corresponding to prior objects that were not detected and to newly detected objects. For the standard model certain pairings cannot co-occur. for instance $i=1,j=2$ and $i=1,j=3$ both include the same prior object. This constraint is encapsulated in an indicator function $A(k\leftrightarrow i,j)$. We do not discuss how to appropriately express or approximate $R_+(k\leftrightarrow i,j)$, which is typically termed the data association step. A variety of approximations have been developed to achieve accurate association at high speed \cite{me_mht, tracking_netflowSSP, mht_blackman, tracking_spline}.

\section{Tracking with Persistent Detector Failures}
As discussed in section \ref{subsec_smm}, the standard tracking model assumes that each object generates a single detection with a certain probability at each timestep, and that the probability of generating a detection is independent both between objects and across timesteps. It also assumes that false detections occur independently of present objects and other false detections, usually following a Poisson process.

Neither of these assumptions are realistic for frequent updates from complex object detectors. The reality is that detections at nearby timesteps are highly correlated. However, handling all potentially correlated detectors simultaneously would require a complex multi-detector data association step, and may not be accurately approximable in real time. This section introduces more efficient methods to incorporate failure persistence into the standard tracking model.

\subsection{False Negative Persistence through Detectability}
As argued in the introduction, the probability of successfully detecting a present object will be correlated across time because that probability is primarily determined by latent features of the object and environment. This suggests that the correlation could be corrected by an augmentation of the object state. Say each object's state includes a binary feature ``detectable". This feature's only impact is to alter the probability of detection as such:
\begin{equation}
r_D(x) \leftarrow \left\{ \begin{array}{lr} r_D(x) & x\text{ detectable} \\ 0 & x\text{ not detectable} \end{array} \right\}
\end{equation}
The detectability, or the probability that an object is detectable, can then be stored and updated as part of each object's state distribution. Specifically, the update expressions \eqref{eq_objmisslik} and \eqref{eq_objmisspost} are modified to include:
\begin{align}
&P_{i,j}(x \text{ detectable} | z_j) = 1 \\
&P_{i,\emptyset}(x \text{ detectable}) = \frac{P_{i,\emptyset}(x \text{ detectable})\left(1 - \int_x r_D(x)\right)}{1 - P_{i,\emptyset}(x \text{ detectable})\int_x r_D(x)}
\end{align}
As the ``detectable" feature is binary, and its actual causes may be too complex to model, it is reasonable to model its change over time as a discrete Markov chain. The original formulation of independent false negatives can be achieved by setting the Markov transition to stationary - that is, the detectability at each new timestep is the same regardless of the previous timestep. The detectability of the undetected objects term $P_U(x)$ can be handled similarly - as this term represents undetected objects, its detectability will be as low as possible at every timestep. New objects $P_{\text{new}}$ can be set to this low detectability for homogeneity, or set to the steady-state detectability value.

\begin{figure}
\centering
\includegraphics[width=3.4in]{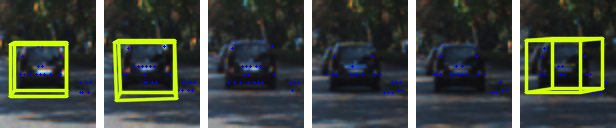}
\caption{Six timesteps from scene 4 of the Kitti tracking dataset, showing a distant leading vehicle that is undetected for three timesteps. Lidar point returns are overlaid in blue, and VoxelJones detections are displayed as 3D box frames.}
\label{fig_examplemissdet}
\end{figure}

An example from the Kitti dataset is shown in Figure \ref{fig_examplemissdet}, in which a tracked vehicle that has been detected for some time is lost for three timesteps then detected again (the detector in question is discussed in section \ref{subsec_voxeljones}). This detection failure is likely due to the distance of the car as well as absorption of lidar scans by black objects. Assume that the object was tracked for some time and its probability of existence is $R=0.999$ (ignoring multi-hypothesis dependencies, survival, etc.). Were detection failures assumed to occur independently with a likelihood of $1-P_D(x) = 0.05$, the probability of object existence by the final timestep would be around $\frac{\left( 1-\int_x P_D(x)P(x) \right)^3 R}{\left( 1-\int_x P_D(x) P(x) \right)^3 R + 1-R} =0.11$. Alternatively, if objects have a steady-state detectability of 0.95 with a transition half-life of a single timestep, the final probability of existence would be 0.875. A higher existence probability increases the likelihood that the detection in the fifth timestep is associated with the tracked object, rather than considered a previously undetected object. The detectability-augmented tracker also assigns a lower probability to detection failure for steadily detected objects - this means that unlikely detections caused by sudden object motion are more likely to be correctly associated.

We are not aware of anyone who has incorporated detection correlation into a probabilistic tracker, despite the simplicity of doing so. Similar terms such as `perceivability' have been used but with different intent \cite{perceivability}. The closest tracker is \cite{learntotrack}, which similarly categorizes objects as 'Tracked' and 'Lost' for periods of time. However, this categorization is performed by a learned algorithm rather than considered part of the probabilistic model, and therefore absolute at each timestep. Non-probabilistic trackers from the computer vision community often assume that objects may go undetected for a contiguous window of time, and initially create time segments of well-tracked objects called tracklets \cite{intro_visiontracking}. Tracklets that are likely to correspond to a single object are then combined, and the object's position in undetected periods can be imputed. This approach to tracking does not inherently reason about the motion of currently-undetected objects, as may be necessary for vehicular applications.

\subsection{Tracking Persistent False Positives}
Intuitively, the only way to characterize correlated false positives is to maintain knowledge of previous false positives. This is equivalent to tracking the false positives as well as true objects. We consider each object to potentially belong to one of two classes, ``genuine'' or ``false''. The multi-object motion model \eqref{eq_smotm} can be altered such that there are two types of objects that enter the scene.
\begin{equation}
    P_{\text{new}}(x) = \left\{ \begin{array}{lr} p_{\text{new,gen.}} P_{\text{new,gen.}}(x) & x \text{ genuine} \\ (1-p_{\text{new,gen.}}) P_{\text{new,false}}(x) & x \text{ false}
    \end{array} \right\}
\end{equation}
Note that rather than tracking two new potential objects, one genuine and one false, the multi-object tracking update tracks a single object using a two-part mixture distribution. Equivalently, two separate tracked objects can be kept, one genuine and the other false. These objects have equivalent data association constraints with other objects, as well as the logical constraint that only one of the two exists. The two-object formulation therefore unnecessarily increases the computational complexity of data association, and so is avoided. The exit of objects is similarly altered.
\begin{align}
r_S(x) = &P(x\text{ genuine}) r_S(x | x \text{ genuine}) \nonumber\\
&+ \left( 1- P(x\text{ genuine})\right) r_S(x | x \text{ false})
\end{align}
In addition, to these alterations, the detection model can be changed if false objects are expected to have different detection properties than genuine ones. Many object detectors, especially learned models, provide a numeric score for the confidence that they have detected a real object. This score is an important tool to differentiate between genuine and false detections, and can easily be incorporated to differentiate between genuine and false measurements.

Finally, the multi-object measurement model can be fundamentally altered by removing the false positive measurements. This is not a loss of capability: consider designing the model such that
\begin{align*}
&\kappa_{\text{new}} p_{\text{new,gen}} = \kappa_F \\
&\int_x P(z|x) P_{\text{new,false}}(x) = P_F(z) \\
&r_S(x | x \text{ false}) = 0
\end{align*}
As false objects appear with a Poisson process independently every timestep and don't persist, this model is equivalent to the original measurement model with poisson-generated false measurements. Thus as with detectability, the degree of persistence can be roughly controlled by the false survival probability $r_S$. For instance, in a simple single-object case where survival probabilities are constant, the overall probability of existence for an object will gradually decrease but the relative probability of genuity will increase.
\begin{align}
    P(x^+ \text{ exists}) = P(x \text{ exists})( &P(x\text{ genuine}) r_S(x\text{ genuine}) + \nonumber \\
    &P(x\text{ false}) r_S(x\text{ false})) \\
    P(x^+ \text{ genuine}) = &\frac{P(x\text{ genuine}) r_S(x\text{ genuine})}{P(x^+ \text{ exists})}
\end{align}

The primary benefit of tracking false objects is that the false detections at any time frame are less likely to be erroneously associated with true objects. A secondary benefit is the ability to better distinguish between true and false detections by leveraging information across time. The detector only accesses a single instant of information and thus cannot make these distinctions. For 3D object detectors on traffic scenes, we make two simple assumptions: false detections are unlikely to move and unlikely to persist when viewed from different angles. These characteristics are factors of the latent state of the object, rather than measurements. Thus they can be incorporated into the multi-object propagation step, either by modifying the single-object motion model or the survival probability $r_{\text{survive}}(x)$. We adopt the latter as it is simpler to handle without splitting each tracked object into a separate genuine and false object.

\begin{figure}
\centering
\includegraphics[width=3.4in]{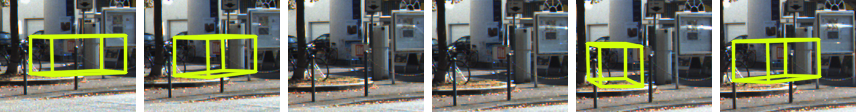}
\caption{Six timesteps from scene 0 of the Kitti tracking dataset, showing false vehicle detections to the side of the road. Detections are displayed as 3D box frames.}
\label{fig_examplefalsedet}
\end{figure}

\begin{figure*}
\centering
\includegraphics[width=6.4in]{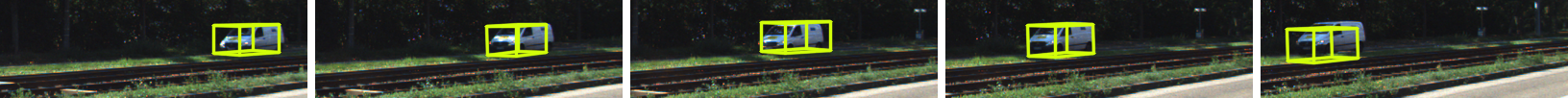}
\caption{Five timesteps from scene 9 of the Kitti tracking dataset, showing low-confidence detections of a moving vehicle. Detections are displayed as 3D box frames.}
\label{fig_exampletruedetfast}
\end{figure*}

Figure \ref{fig_examplefalsedet} shows a segment of the Kitti dataset in which the detector of Section \ref{subsec_voxeljones} generates a false detection near the road. The cause seems to be a combination of objects such as a bicycle, signpost, and barrier poles. The detector's confidence score is around 0.15,  meaning each detection has a relative probability of 0.15 of being legitimate as opposed to false. The actual tracking result is a function of motion model and measurement model parameters, so we instead set a simple hypothetical example: the probability of false detections or undetected objects is 0.01 of the probability of the each detection originating from an object in the same spot. If false detections are tracked, this one will be maintained until it disappears a few timesteps later (as the viewing perspective changes). If false detections are not tracked, this set of detections will be considered either a sequence of independent false detections or a detected object. The tracked object corresponding to these detections will have existence probability $\frac{.15^4}{.15^4 + .01^3} = .998$. Figure \ref{fig_exampletruedetfast} shows another segment with a sequence of low-confidence detections, in this case correctly corresponding to a vehicle. In this case, the tracked object is quickly determined to be moving at over 6 m/s and its probability of genuity is increased as a result.

The concept of tracking false detections has been implicitly performed by some trackers. The vehicle tracker in \cite{simpletracking1} tracks all likely detections, but only reports tracked objects whose associated detections have high scores. The use of temporal information to distinguish between true and false detections has also been performed in a limited fashion: some visual trackers separate relevant objects from background environment based solely on motion \cite{intro_visiontracking}. However, as with persistent missed detections, these techniques were not to our knowledge previously incorporated into the probabilistic model for multi-object tracking.
\section{Vehicle tracking implementation}
A hypothesis-oriented multiple-hypothesis tracker is applied to vehicle tracking. Vehicles are considered to occupy 2D rectangles flat along the ground, with height and roll/pitch considered unimportant for sensing or tracking. This is often termed a bird's-eye-view (BEV) representation. Vehicles are parameterized by position, orientation, length and width, speed in the direction of orientation, and rate of angle change. The sensing vehicle's motion is assumed to be known accurately, so objects are tracked in absolute positions rather than positions relative to the vehicle. Each detection provides position, orientation, and shape estimates in addition to a score corresponding to the expected probability that this is a genuine detection. While persistent false detections and missed detections are handled as discussed, errors in the detections' estimated values are assumed to be independent across time, objects, and features. In other words, the detector error is assumed to be white noise. This assumption is not necessarily more realistic than that of independent failures, and could be addressed by state augmentation \cite{sensorbiasaugment}. However, short-term errors in position or shape estimation are not considered as significant as detection failures.

A 2D grid of 3-meter square tiles is maintained by the tracker for several functions. The ground surface of each tile is estimated and used to convert between BEV, 3D, and image-space positions. Line-of-sight sensor occlusion is determined for each tile based on the lidar returns, and used to estimate occlusion probability for tracked objects. An occupancy grid \cite{intro_gridmain} is used to model the expected rate and distribution of untracked vehicles entering the system ($\kappa_U$ and $P_U(x)$ in section \ref{subsec_backmot}). The occupancy grid is updated by local mixing at every timestep, simulating the motion of untracked objects and the entry of new objects from outside the field of view. Finally, the importance of performing detection on each tile at each time can be roughly estimated \cite{williams2007information}. This is used to speed up a detector, as discussed in Section \ref{subsec_resultsvj}.

As a complete specification of the tracker would be much longer, the code itself is available at github.com/motrom/kittitracking-pdfmht for reproduction or comparison purposes.

\subsection{Alternative Tracking Methods}
Much of vehicular tracking research has focused on tracking objects within a camera's view (also called 2D tracking), despite 3D information being valuable for many vehicular applications. The two domains are not equivalent as an estimated position in bird's-eye-view may not be as accurate when translated to camera view and vice versa. The method AB3DMOT \cite{simpletracking1} has code available online and was previously tested using the strong detector discussed in the next section. Thus this method is used for comparison. Other methods such as \cite{tracking_pmbm, fakes_kampker, complexeryolo, robustmmmot, trivedi360tracker} do not have public implementations. \cite{tracking_pmbm, fakes_kampker, complexeryolo} present probabilistic trackers of similar structure to ours. AB3DMOT follows simple rules for handling object existence and detector failure, and \cite{robustmmmot, trivedi360tracker} learn tracker rules from example data.

\subsection{Detectors}
\subsubsection{Strong Detector}
Point-RCNN \cite{pointrcnn}, a deep CNN lidar-based detector, was chosen to represent state-of-the-art vehicle detectors. It was trained on the Kitti detection training set, leaving out samples that correspond to the first 10 scenes of the Kitti tracking dataset. Those scenes are used to validate the tracker.

\subsubsection{Lightweight Lidar Detector}
\label{subsec_voxeljones}
If current autonomous vehicle object detectors fall short anywhere, it is in interpretability and computational demand. We propose a simple Viola-Jones-type object detector for vehicular lidar, which we refer to as VoxelJones. One of the challenges for classifying on lidar data is deriving useful features from sets of 3D points. Decision trees make binary splits rather than algebraic transformations, so features are not required to be numeric. We select a simple but intuitive and numerous set of binary features. The box of -3 to 3 meters lengthwise with respect to an object's center, -2 to 2 meters widthwise, and .125 to 2.625 meters above the ground is discretized into $0.125$-meter binary voxels (3D cubes). A voxel is positive if at least one lidar point falls within that region of space. Any box-shaped union of voxels may be used as a feature in the classifier.

We use second-order gradient boosting to train weighted regression trees as in \cite{fakes_xgboost}. The decision trees are of depth 3 and so each contain 7 splits and 8 resulting values. A first classifier of 10 trees operates on anchors spaced every 0.5 meters and $\frac{\pi}{8}$ radians. The attentional cascade eliminates 99.8\% of negative inputs, while erroneously eliminating 14\% of positive inputs.
Positively classified areas are split into anchors spaced at 0.125 meters and $\frac{\pi}{16}$ radians, and further classified with 20 more trees. As Kitti's vehicle class does not include large vehicles, all detected vehicles are considered to have dimensions of 4m length, 1.76m width, and 1.7m height.

In addition to inference speed, decision trees and box features have the advantage of interpretability. Appendix \ref{appendix_interpret} includes visual interpretations of the algorithm's parameters and of example input.

The algorithm was trained with the Kitti object detection training dataset, leaving out samples corresponding to the utilized tracking dataset. Its performance on the Kitti object detection test benchmark is shown in Table \ref{tab_voxeljonesresults}. It is compared to deep network detections from the Kitti public leaderboard. The cutoff metric is 0.7 BEV overlap. Algorithm speed is also reported in terms of average detections per second. For most methods, including ours, there is little fluctuation from the average.
\begin{table}[ht]
\centering
\renewcommand{\arraystretch}{1.3}
\caption{Detector Performance on Kitti BEV Benchmark}
\label{tab_voxeljonesresults}
\begin{tabular}{|c|c|c|c|c|c|}
\hline \, & AP Easy & AP Moderate & AP Hard & FPS & GPU \\
\hline VoxelJones & 65.3 & 54.5 & 49.9 & 6 & N \\
\hline \cite{detect_3DFCN} & 69.4 & 62.5 & 55.9 & 1 & Y \\
\hline \cite{pointrcnn} & 89.5 & 85.7 & 79.1 & 10 & Y \\
\hline \cite{fakes_dnlidarcpu} & -- & -- & 15.4 & 10 & N \\
\hline
\end{tabular}
\end{table}
Our detection method achieves similar performance to the early deep network \cite{detect_3DFCN}, and superior performance to the only other public method that can be applied in real-time on a CPU \cite{fakes_dnlidarcpu}. However, state-of-the-art detectors are notably better.

\subsubsection{Detector Fusion}
Recent computer vision research has led to object detectors that estimate 3D position from a single image. However, the distance estimation of these methods are still inaccurate compared to stereo or rotational lidar. We track using a mono-camera deep detector MonoGRNet \cite{monogrnet} (trained on the same data as the other detectors) as well as sparsified lidar data.

Six of the sixty-four lasers from the Kitti dataset's lidar are used. The cost of a lidar sensor scales with its resolution, and cameras are inexpensive sensors so the combination of the two provides a low-cost approach to object detection. However, at this resolution it is difficult to distinguish between vehicles and other objects using lidar data alone, so a learned model is foregone in favor of a simple segmentation of each laser's returns. A contiguous set of lidar returns with less than .4 meters distance between each is considered a segment. Segments that are long enough and of the right shape to be a vehicle are treated as full detections (position, orientation, and shape), while other segments are treated as inaccurate measurements of the vehicle position.

\begin{figure*}[ht]
    \centering
    \includegraphics[width=7.0in]{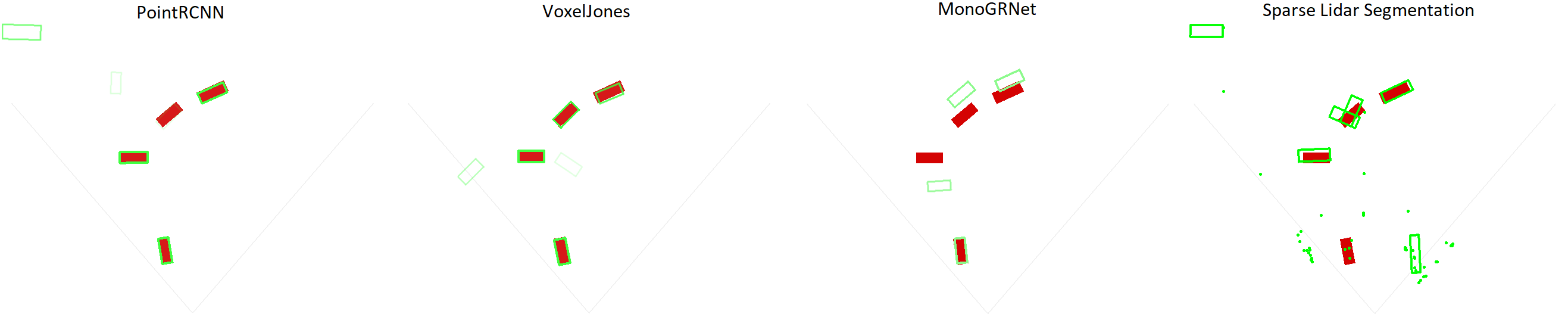}
    \includegraphics[width=4.5in]{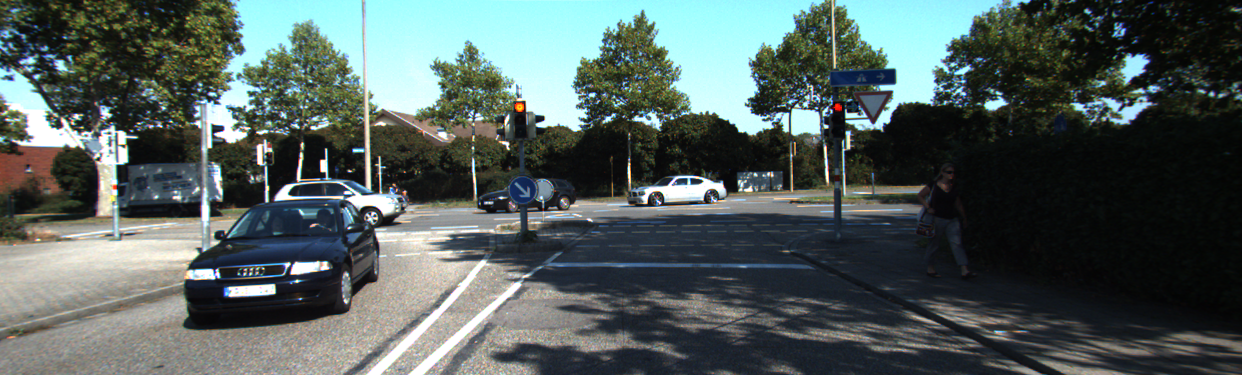}
    \caption{Scene 1 sample 368 of the kitti tracking training dataset. Annotated vehicles are shown as red boxes. Detections are in green and shaded by detector confidence. Lidar segments that are too small to determine orientation or size are shown as points instead.}
    \label{fig_detectorsimg}
\end{figure*}

Figure \ref{fig_detectorsimg} shows a single sample and the estimates from each of the detectors. Point-RCNN misses one vehicle due to occlusion and reports one false detection with very low confidence (note that detection in the far left is actually an unannotated vehicle). VoxelJones actually detects the occluded vehicle but reports several false detections. MonoGRNet correctly detects all vehicles but with inaccurate positions. The segmentation of sparse lidar points detects each vehicle at least once, but also detects many irrelevant objects.

\section{Tracking Performance}
The tracker is tested on the first 10 scenes from the Kitti tracking training dataset. This subset contains 281 unique vehicles with 8307 total annotations. The proposed tracker is denoted as PDFMHT (Persistent Detector Failures in Multiple Hypothesis Tracking) in the following results.

\subsection{Tracking on Strong Detector}
The tracker is first applied to detections from Point-RCNN \cite{pointrcnn}.

\begin{figure*}[ht]
\centering
\includegraphics[width=6.7in]{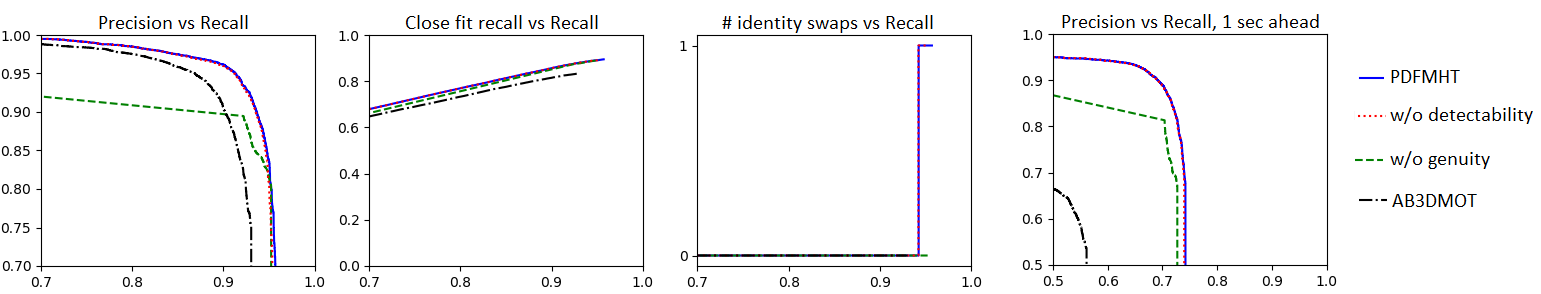}
\caption{Performance of trackers on the Kitti tracking subset using the Point-RCNN detector.}
\label{fig_strongtrackingresults}
\end{figure*}

Common MOT performance metrics used in \cite{data_mot, data_kitti} calculate false positive, false negative, and identity inconsistency errors from a single output estimate from the tracker, then compile these metrics into several aggregate scores. As mentioned by \cite{simpletracking1}, such metrics do not consider that these errors may not have equal importance for a given application, and that most trackers can be adapted to report a higher or lower number of estimates to prioritize completeness or precision respectively. Additionally, the ability of the tracker to determine the latent features of tracked objects is important for robotics applications. Figure \ref{fig_strongtrackingresults} presents tracker performance in four plots. The horizontal axis of each plot corresponds to recall, or the fraction of annotated vehicles that were detected. An estimate is considered to correctly match an annotation if the BEV overlap (intersection over union) between the two exceeds 0.3. The vertical axis of the leftmost plot displays precision, or the fraction of reported estimates that were correctly matched. The second plot displays recall under a stricter overlap requirement of 0.7, as is used in the Kitti detection benchmark. The vertical axis of the third plot reports the number of identity switches, as defined by the CLEAR MOT metrics. The final plot is also a precision-recall plot, but calculated for estimates one second ahead using the tracker's motion model as a predictor \footnote{For both trackers, the sensing vehicle's motion is directly accounted for rather than predicted.}. The overlap requirement is lowered to 0.1 for this plot, as motion prediction is challenging without utilizing context or more powerful prediction models. We also report single values of CLEAR MOTA and MOTP in Table \ref{tab_trackingresultsT}, using the 0.3 overlap requirement.

The rule-based decisions for including and removing object estimates in AB3DMOT \cite{simpletracking1} are shown to perform poorly compared to the probabilistic model used in PDFMHT, even when detector failures are not considered. However, other tracker details such as occlusion and occupancy reasoning may contribute heavily to this difference in performance. Utilizing genuity in the tracker improves performance substantially - the leftmost plot of Figure \ref{fig_strongtrackingresults} reveals that the traditional tracker model cannot distinguish between true and false objects beyond a certain level of confidence. For high-recall low-precision settings where even highly uncertain estimates are reported, genuity makes less of an impact. The addition of a detectability feature is not shown to impact this tracker and detector. As low-confidence detections can be safely included with the genuity modification, and occlusion is explicitly accounted for, there may be few unexpected missed detections.

\begin{table}[ht]
\centering
\renewcommand{\arraystretch}{1.3}
\caption{Tracker Performance, CLEAR MOTA Metric @ .3 BEV IoU}
\label{tab_trackingresultsT}
\begin{tabular}{|c|c|c|}
\hline Detector & Tracker & MOTA \\
\hline PointRCNN & PDFMHT & 86.6 \\
\hline \, & w/o Detectability & 86.4 \\
\hline \, & w/o Genuity & 81.4 \\
\hline \, & AB3DMOT & 82.5 \\
\hline Voxel Jones & PDFMHT & 74.7 \\
\hline \, & w/o Detectability & 73.4 \\
\hline \, & w/o Genuity & 60.6 \\
\hline \, & subselect environment & 72.4  \\
\hline MonoGRNet & PDFMHT & 42.6  \\
\hline MonoGRNet + sparse lidar & PDFMHT & 46.5 \\
\hline \, & w/o Genuity & 33.4 \\
\hline \, & w/o Detectability & 45.7 \\
\hline
\end{tabular}
\end{table}

\subsection{Tracking on Lightweight Lidar Detector} \label{subsec_resultsvj}
\begin{figure*}[ht]
\centering
\includegraphics[width=7.0in]{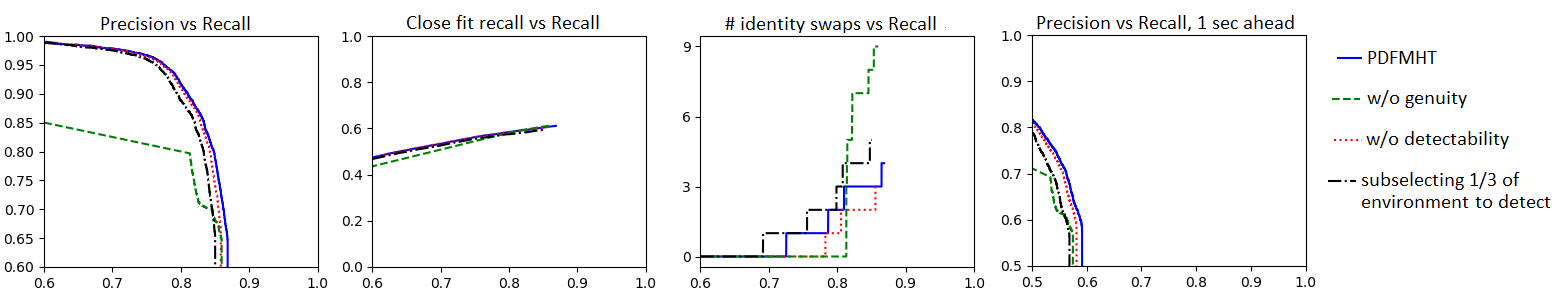}
\caption{Performance of trackers on the Kitti tracking subset using the VoxelJones detector.}
\label{fig_vjtrackingresults}
\end{figure*}

As in the previous subsection, we plot several metrics in Figure \ref{fig_vjtrackingresults} and report standard MOTA in Table \ref{tab_trackingresultsT}. The tracker is compared to alternate versions without false positive or false negative persistence modeled, as well as against a simple track-before-detect experiment. The tracker is used to subselect a third of the viewed area at every timestep on which to apply the detector. This more than doubles the detector's speed, with the sublinear improvement explained by costly preprocessing and the fact that the selected regions are less likely to be removed in the attentional cascade.

The tracker's performance is again highly improved by adding persistent false object reasoning. This is reasonable as the object detector used provides many false positives. Detectability improves performance in a few instances, unlike for the strong detector. Additionally, the tracker can be used to increase the detector's speed with only a minor performance penalty. The subselected VoxelJones+tracker pair runs at over 10 updates per second. Interestingly, for 1-second predictions the accuracy of this tracker exceeds that of the AB3DMOT tracker using state-of-the-art detections. Again, this difference in performance can not solely be attributed to the persistent failure reasoning, but it shows the value of an accurate tracker compared to solely an accurate detector.

\subsection{Tracking with Fused Detectors}
PDFMHT is modified slightly to handle these complementary sensors. A steady-state bias feature is maintained for each object, which stores the longitudinal bias of MonoGRNet detections. In the lateral direction, the detector is substantially more accurate. Each laser is treated as a separate sensor and handled with a separate update step. This approach maintains the one-to-one restriction of the standard tracking model and enables simple occlusion reasoning. Finally, the approach for genuity is changed for lidar segments. Rather than track a large number of false lidar objects, lidar-only objects are added to the undetected estimates, which in PDFMHT are stored in an occupancy grid. This still achieves the desired resilience to repeated lidar detections in the same region, as they will be associated with and then reassigned to the undetected objects each update step. Any object associated with a camera measurement is handled like the previously discussed genuity estimates.

\begin{figure*}[ht]
\centering
\includegraphics[width=5.5in]{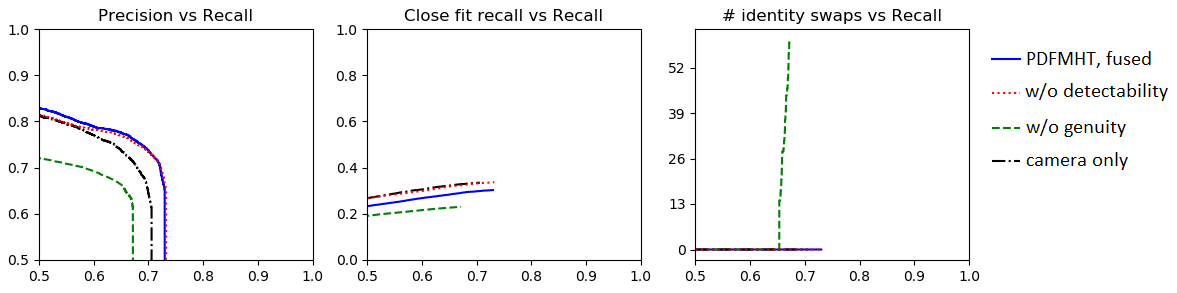}
\caption{Performance of trackers on the Kitti tracking subset using the MonoGRNet detector  as well as segment detections from sparse lidar.}
\label{fig_fusetrackingresults}
\end{figure*}

Figure \ref{fig_fusetrackingresults} and Table \ref{tab_trackingresultsT} show results of the fused tracking model. Note that for the genuity-less model, we still do not track lidar-only objects. Otherwise, a large number of non-vehicle objects would be reported. However, without maintaining the occupancy grid of false measurements, lidar detections are often matched to camera detections when they more likely correspond to previously detected non-vehicle objects. This effect is extreme enough to make a fused tracker less accurate than a tracker using only the MonoGRNet detections. With genuity reasoning, the fusion of detector models is shown to improve upon the MonoGRNet-only tracker.

\section{Conclusion}
We present modifications to the standard probabilistic multi-object tracking formulation that more realistically capture some common properties of robotic object detectors. Objects that are persistently undetected for a short time can be handled by maintaining an estimate of object detectability. False detections that persist in a certain region can be handled by tracking potentially false objects and using a combination of detector confidence and latent features to distinguish between genuine and false tracked objects. A 3D vehicle tracker was implemented using a standard probabilistic model with these modifications and tested on a public dataset. Persistence-aware tracking is shown to be valuable for further improving the detection performance of state-of-the-art deep networks, for enabling real-time detection with a classic detection algorithm without a graphics processing unit, and for correctly fusing detections from complementary sensors.


%

\appendices

\section{Lidar Object Detector Interpretability} \label{appendix_interpret}
This section covers two ways in which the VoxelJones object detector can be interpreted visually.
\subsection{Visualizing the model}
\begin{figure}[ht]
\centering
\includegraphics[width=3.2in]{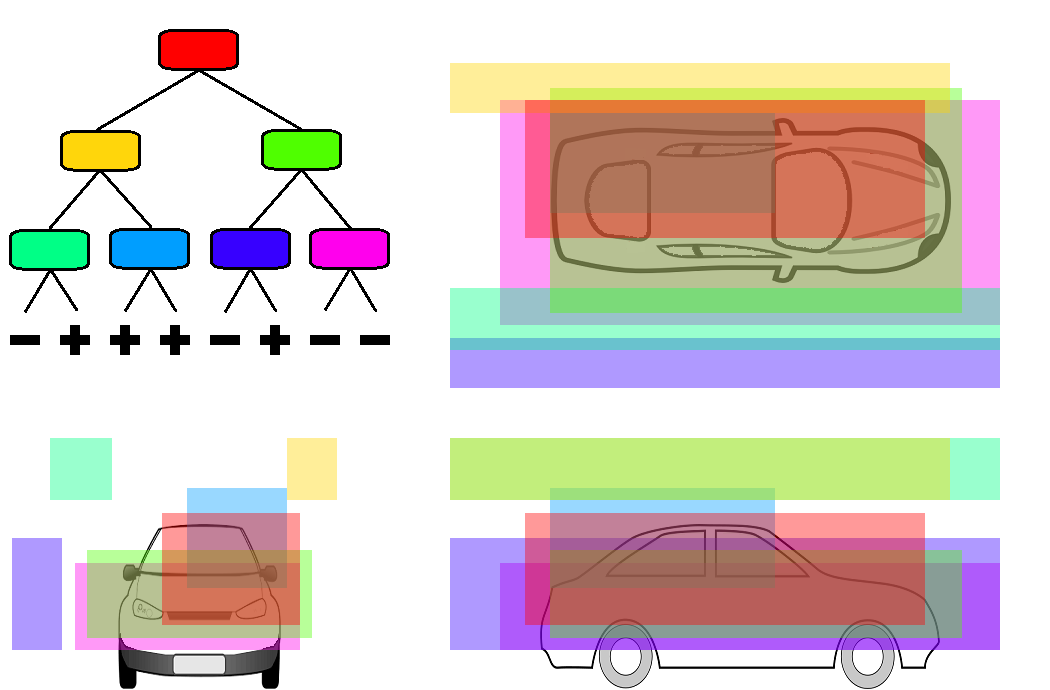}
\caption{Visualization of box splits for the first tree of the trained VoxelJones object detector.}
\label{fig_appendixvjtree}
\end{figure}

As all features for the detector are simply point checks within 3D boxes, they can be visualized as 3D shapes. Figure \ref{fig_appendixvjtree} visualizes all feature boxes from the first tree of our trained model from three perspectives. For each colored box, a split to the left is taken if there is a lidar point within this box. The final scores for this tree are denoted as positive or negative. The first box (red) checks a broad section of where a car would be. Splits to the left check high regions, presumably to negatively score tall and large objects like buildings and trees. If no lidar detections lie in the first box, the next splits search for detections in nearby areas.

\subsection{Visualizing inputs}
\begin{figure*}
    \centering
    \subfloat[]{
    \includegraphics[width=2.0in]{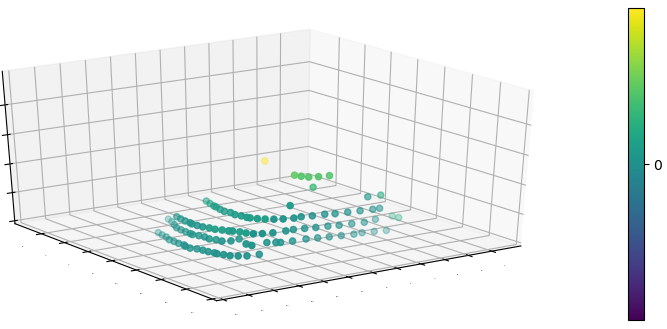}
    }
    \subfloat[]{
    \includegraphics[width=1.2in]{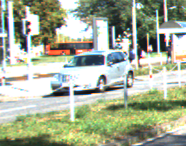}
    }
    \subfloat[]{
    \includegraphics[width=2.0in]{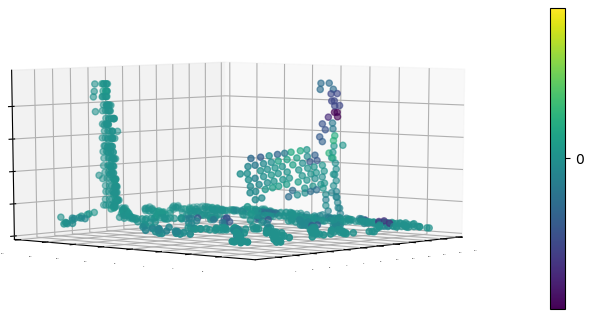}
    }
    \subfloat[]{
    \includegraphics[width=1.2in]{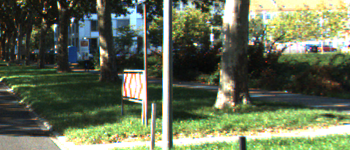}
    }
    \caption{Visualizations of object detector input importance. (a) point cloud and (b) image of a correctly detected vehicle, (c) point cloud and (d) image of a correctly ignored sign. Lighter point colors correspond to positive impact on likelihood of being an object, darker colors negative.}
    \label{fig_appendixvjpoints}
\end{figure*}

The concept of feature importance has a clear mathematical formulation for boosted trees, based on the change in score directly caused by splits on a single feature \cite{appendix_treeimportance}. In our case, the feature space is too broad for this metric to be easily interpretable, but a similar approach can be taken for an input set of lidar points. The points that lie inside a feature box are considered to have importance correlated to the score of that tree, divided by the number of similar points. Points or small sets of points that uniquely contributed to positive or negative classification can thus be located. Figure \ref{fig_appendixvjpoints} shows an example of a true detection and false detection with input point importances displayed by color. A single lidar point in the true example hits the roof of the far side of the car and strongly impacts the classification. The most significant points in the false example lie on the thin pole, and curiously on some patches of ground. This could imply the model has learned from the data to expect vehicles on flat surfaces.


\ifCLASSOPTIONcaptionsoff
  \newpage
\fi



\bibliographystyle{IEEEtran}
\bibliography{./bibtex/bib/refs}

\begin{thebibliography}{10}
\providecommand{\url}[1]{#1}
\csname url@samestyle\endcsname
\providecommand{\newblock}{\relax}
\providecommand{\bibinfo}[2]{#2}
\providecommand{\BIBentrySTDinterwordspacing}{\spaceskip=0pt\relax}
\providecommand{\BIBentryALTinterwordstretchfactor}{4}
\providecommand{\BIBentryALTinterwordspacing}{\spaceskip=\fontdimen2\font plus
\BIBentryALTinterwordstretchfactor\fontdimen3\font minus
  \fontdimen4\font\relax}
\providecommand{\BIBforeignlanguage}[2]{{%
\expandafter\ifx\csname l@#1\endcsname\relax
\typeout{** WARNING: IEEEtran.bst: No hyphenation pattern has been}%
\typeout{** loaded for the language `#1'. Using the pattern for}%
\typeout{** the default language instead.}%
\else
\language=\csname l@#1\endcsname
\fi
#2}}
\providecommand{\BIBdecl}{\relax}
\BIBdecl

\bibitem{intro_lazytracking}
E.~Bochinski, V.~Eiselein, and T.~Sikora, ``High-speed tracking-by-detection
  without using image information,'' in \emph{2017 14th IEEE International
  Conference on Advanced Video and Signal Based Surveillance (AVSS)}.\hskip 1em
  plus 0.5em minus 0.4em\relax IEEE, 2017, pp. 1--6.

\bibitem{fakes_simpletracking2}
G.~G{\"u}nd{\"u}z and T.~Acarman, ``A lightweight online multiple object
  vehicle tracking method,'' in \emph{2018 IEEE Intelligent Vehicles Symposium
  (IV)}.\hskip 1em plus 0.5em minus 0.4em\relax IEEE, 2018, pp. 427--432.

\bibitem{simpletracking1}
\BIBentryALTinterwordspacing
X.~Weng and K.~Kitani, ``{A Baseline for 3D Multi-Object Tracking},''
  \emph{arXiv:1907.03961}, 2019. [Online]. Available:
  \url{https://arxiv.org/pdf/1907.03961.pdf}
\BIBentrySTDinterwordspacing

\bibitem{me_mwo}
M.~Motro and J.~Ghosh, ``Measurement-wise occlusion in multi-object tracking,''
  in \emph{2018 21st International Conference on Information Fusion
  (FUSION)}.\hskip 1em plus 0.5em minus 0.4em\relax IEEE, 2018, pp. 2384--2391.

\bibitem{localizationandtracking}
F.~Moosmann and C.~Stiller, ``Joint self-localization and tracking of generic
  objects in 3d range data,'' in \emph{2013 IEEE International Conference on
  Robotics and Automation}.\hskip 1em plus 0.5em minus 0.4em\relax IEEE, 2013,
  pp. 1146--1152.

\bibitem{mwo_variationalradar}
A.~Scheel and K.~Dietmayer, ``Tracking multiple vehicles using a variational
  radar model,'' \emph{arXiv preprint arXiv:1711.03799}, 2017.

\bibitem{data_recentsurvey}
Y.~Kang, H.~Yin, and C.~Berger, ``Test your self-driving algorithm: An overview
  of publicly available driving datasets and virtual testing environments,''
  \emph{IEEE Transactions on Intelligent Vehicles}, vol.~4, no.~2, pp.
  171--185, 2019.

\bibitem{data_mot}
\BIBentryALTinterwordspacing
A.~Milan, L.~Leal{-}Taix{\'{e}}, I.~D. Reid, S.~Roth, and K.~Schindler,
  ``{MOT16:} {A} benchmark for multi-object tracking,'' \emph{CoRR}, vol.
  abs/1603.00831, 2016. [Online]. Available:
  \url{http://arxiv.org/abs/1603.00831}
\BIBentrySTDinterwordspacing

\bibitem{fakes_violajones}
Y.-Q. Wang, ``An analysis of the viola-jones face detection algorithm,''
  \emph{Image Processing On Line}, vol.~4, pp. 128--148, 2014.

\bibitem{detect_pointpillars}
A.~H. Lang, S.~Vora, H.~Caesar, L.~Zhou, J.~Yang, and O.~Beijbom,
  ``{PointPillars}: Fast encoders for object detection from point clouds,''
  Dec. 2018.

\bibitem{monogrnet}
Z.~Qin, J.~Wang, and Y.~Lu, ``Monogrnet: A geometric reasoning network for
  monocular 3d object localization,'' in \emph{Proceedings of the AAAI
  Conference on Artificial Intelligence}, vol.~33, 2019, pp. 8851--8858.

\bibitem{traffic_monopmbm}
S.~Scheidegger, J.~Benjaminsson, E.~Rosenberg, A.~Krishnan, and
  K.~Granstr{\"o}m, ``Mono-camera 3d multi-object tracking using deep learning
  detections and pmbm filtering,'' in \emph{2018 IEEE Intelligent Vehicles
  Symposium (IV)}.\hskip 1em plus 0.5em minus 0.4em\relax IEEE, 2018, pp.
  433--440.

\bibitem{fakes_shallowlidarclassify}
M.~Kusenbach, M.~Himmelsbach, and H.~Wuensche, ``A new geometric {3D} {LiDAR}
  feature for model creation and classification of moving objects,'' in
  \emph{2016 {IEEE} Intelligent Vehicles Symposium ({IV})}, Jun. 2016, pp.
  272--278.

\bibitem{fakes_shallowlidarclass2}
J.~Behley, V.~Steinhage, and A.~B. Cremers, ``Laser-based segment
  classification using a mixture of bag-of-words,'' in \emph{2013 {IEEE/RSJ}
  International Conference on Intelligent Robots and Systems}, Nov. 2013, pp.
  4195--4200.

\bibitem{fakes_kampker}
A.~Kampker, M.~Sefati, A.~S.~A. Rachman, K.~Kreisk{\"o}ther, and P.~Campoy,
  ``Towards multi-object detection and tracking in urban scenario under
  uncertainties.'' in \emph{VEHITS}, 2018, pp. 156--167.

\bibitem{fakes_shallowlidarrules2}
A.~Asvadi, C.~Premebida, P.~Peixoto, and U.~Nunes, ``{3D} lidar-based static
  and moving obstacle detection in driving environments: An approach based on
  voxels and multi-region ground planes,'' \emph{Rob. Auton. Syst.}, vol.~83,
  pp. 299--311, Sep. 2016.

\bibitem{data_kitti}
A.~Geiger, P.~Lenz, C.~Stiller, and R.~Urtasun, ``Vision meets robotics: The
  kitti dataset,'' \emph{The International Journal of Robotics Research},
  vol.~32, no.~11, pp. 1231--1237, 2013.

\bibitem{fakes_pydriver}
L.~Plotkin, ``Pydriver: Entwicklung eines frameworks f{\"u}r r{\"a}umliche
  detektion und klassifikation von objekten in fahrzeugumgebung,'' Ph.D.
  dissertation, Bachelor's Thesis, Karlsruhe Institute of Technology,
  Karlsruhe, Germany, 2015.

\bibitem{fakes_second}
Y.~Yan, Y.~Mao, and B.~Li, ``{SECOND}: Sparsely embedded convolutional
  detection,'' \emph{Sensors}, vol.~18, no.~10, Oct. 2018.

\bibitem{tracking_pmbm}
{\'A}.~F. Garc{\'\i}a-Fern{\'a}ndez, J.~L. Williams, K.~Granstrom, and
  L.~Svensson, ``Poisson multi-bernoulli mixture filter: direct derivation and
  implementation,'' \emph{IEEE Transactions on Aerospace and Electronic
  Systems}, 2018.

\bibitem{mwo_multimeasurement}
C.~Adam, R.~Schubert, and G.~Wanielik, ``Radar-based extended object tracking
  under clutter using generalized probabilistic data association,'' in
  \emph{16th International {IEEE} Conference on Intelligent Transportation
  Systems ({ITSC} 2013)}, Oct. 2013, pp. 1408--1415.

\bibitem{me_mht}
M.~Motro and J.~Ghosh, ``Scaling data association for hypothesis-oriented
  mht,'' in \emph{2019 22nd International Conference on Information Fusion
  (FUSION)}.\hskip 1em plus 0.5em minus 0.4em\relax IEEE, 2019.

\bibitem{tracking_netflowSSP}
P.~Lenz, A.~Geiger, and R.~Urtasun, ``{FollowMe}: Efficient online min-cost
  flow tracking with bounded memory and computation,'' in \emph{Proceedings of
  the {IEEE} International Conference on Computer Vision}.\hskip 1em plus 0.5em
  minus 0.4em\relax cv-foundation.org, 2015, pp. 4364--4372.

\bibitem{mht_blackman}
S.~S. Blackman, ``Multiple hypothesis tracking for multiple target tracking,''
  \emph{IEEE Aerospace and Electronic Systems Magazine}, vol.~19, no.~1, pp.
  5--18, 2004.

\bibitem{tracking_spline}
K.~Date and R.~Nagi, ``Tracking multiple maneuvering targets using integer
  programming and spline interpolation,'' in \emph{2018 21st International
  Conference on Information Fusion (FUSION)}.\hskip 1em plus 0.5em minus
  0.4em\relax IEEE, 2018, pp. 1293--1300.

\bibitem{perceivability}
{Ning Li} and X.~R. Li, ``Target perceivability and its applications,''
  \emph{IEEE Trans. Signal Process.}, vol.~49, no.~11, pp. 2588--2604, Nov.
  2001.

\bibitem{learntotrack}
Y.~Xiang, A.~Alahi, and S.~Savarese, ``Learning to track: Online multi-object
  tracking by decision making,'' in \emph{Proceedings of the IEEE international
  conference on computer vision}, 2015, pp. 4705--4713.

\bibitem{intro_visiontracking}
Z.~Wu, N.~I. Hristov, T.~H. Kunz, and M.~Betke, ``Tracking-reconstruction or
  reconstruction-tracking? comparison of two multiple hypothesis tracking
  approaches to interpret 3d object motion from several camera views,'' in
  \emph{2009 Workshop on Motion and Video Computing (WMVC)}.\hskip 1em plus
  0.5em minus 0.4em\relax IEEE, 2009, pp. 1--8.

\bibitem{sensorbiasaugment}
B.~Friedland, ``Treatment of bias in recursive filtering,'' \emph{IEEE
  Transactions on Automatic Control}, vol.~14, no.~4, pp. 359--367, 1969.

\bibitem{intro_gridmain}
M.~Schreier, V.~Willert, and J.~Adamy, ``Compact representation of dynamic
  driving environments for adas by parametric free space and dynamic object
  maps,'' \emph{IEEE Transactions on Intelligent Transportation Systems},
  vol.~17, no.~2, pp. 367--384, 2015.

\bibitem{williams2007information}
J.~L. Williams, ``Information theoretic sensor management,'' Ph.D.
  dissertation, Massachusetts Institute of Technology, 2007.

\bibitem{complexeryolo}
M.~Simon, K.~Amende, A.~Kraus, J.~Honer, T.~Samann, H.~Kaulbersch, S.~Milz, and
  H.~Michael~Gross, ``Complexer-yolo: Real-time 3d object detection and
  tracking on semantic point clouds,'' in \emph{Proceedings of the IEEE
  Conference on Computer Vision and Pattern Recognition Workshops}, 2019, pp.
  0--0.

\bibitem{robustmmmot}
W.~Zhang, H.~Zhou, S.~Sun, Z.~Wang, J.~Shi, and C.~C. Loy, ``Robust
  multi-modality multi-object tracking,'' \emph{arXiv preprint
  arXiv:1909.03850}, 2019.

\bibitem{trivedi360tracker}
A.~Rangesh and M.~M. Trivedi, ``No blind spots: Full-surround multi-object
  tracking for autonomous vehicles using cameras \& lidars,'' \emph{IEEE
  Transactions on Intelligent Vehicles}, 2019.

\bibitem{pointrcnn}
S.~Shi, X.~Wang, and H.~Li, ``Pointrcnn: 3d object proposal generation and
  detection from point cloud,'' in \emph{Proceedings of the IEEE Conference on
  Computer Vision and Pattern Recognition}, 2019, pp. 770--779.

\bibitem{fakes_xgboost}
T.~Chen and C.~Guestrin, ``Xgboost: A scalable tree boosting system,'' in
  \emph{Proceedings of the 22nd acm sigkdd international conference on
  knowledge discovery and data mining}.\hskip 1em plus 0.5em minus 0.4em\relax
  ACM, 2016, pp. 785--794.

\bibitem{detect_3DFCN}
B.~Li, ``3d fully convolutional network for vehicle detection in point cloud,''
  in \emph{2017 IEEE/RSJ International Conference on Intelligent Robots and
  Systems (IROS)}.\hskip 1em plus 0.5em minus 0.4em\relax IEEE, 2017, pp.
  1513--1518.

\bibitem{fakes_dnlidarcpu}
K.~Minemura, H.~Liau, A.~Monrroy, and S.~Kato, ``Lmnet: Real-time multiclass
  object detection on cpu using 3d lidar,'' in \emph{2018 3rd Asia-Pacific
  Conference on Intelligent Robot Systems (ACIRS)}.\hskip 1em plus 0.5em minus
  0.4em\relax IEEE, 2018, pp. 28--34.

\bibitem{appendix_treeimportance}
J.~Friedman, T.~Hastie, and R.~Tibshirani, \emph{The elements of statistical
  learning}.\hskip 1em plus 0.5em minus 0.4em\relax Springer series in
  statistics New York, 2001, vol.~1, no.~10.

\end{thebibliography}
\end{document}